%% file: main.tex
\documentclass[letterpaper, 10 pt, conference]{ieeeconf}
\IEEEoverridecommandlockouts 

\usepackage[utf8]{inputenc}
\usepackage[T1]{fontenc}
\usepackage{amsmath,amsfonts}
\usepackage{amssymb}
\usepackage{mathtools}
\usepackage{booktabs}
\usepackage{array}
\usepackage{multirow} 
\usepackage{makecell}
\usepackage{tabularx}
\usepackage{threeparttable}
\usepackage[table,xcdraw]{xcolor}
\usepackage{float}
\usepackage{siunitx}    %
\usepackage{graphicx}
\usepackage{caption}
\usepackage{subcaption}
\usepackage[font=small,labelfont=bf,tableposition=top]{caption}
\usepackage{stfloats}
\usepackage{adjustbox}
\usepackage[table]{xcolor}

\usepackage{algorithm}
\usepackage{algorithmicx}
\usepackage{algpseudocode}

\usepackage{textcomp}
\usepackage{url}
\usepackage{verbatim}
\usepackage{cite}

\definecolor{rblue}{rgb}{0,0.5,1}
\definecolor{awesome}{rgb}{1.0, 0.13, 0.32}
\definecolor{hollywoodcerise}{rgb}{0.96, 0.0, 0.63}
\definecolor{lasallegreen}{rgb}{0.03, 0.47, 0.19}
\definecolor{hanpurple}{rgb}{0.32, 0.09, 0.98}
\definecolor{green(pigment)}{rgb}{0.0, 0.65, 0.31}

\definecolor{bestbg}{HTML}{DDF7E3}
\definecolor{secondbg}{HTML}{FFE9E9}

\definecolor{grpName}{HTML}{F0F7FF}
\definecolor{grpMask}{HTML}{FFF7EC}
\makeatletter
\let\NAT@parse\undefined
\makeatother
\usepackage[pagebackref=false,breaklinks=true,colorlinks,bookmarks=false]{hyperref}
\hypersetup{colorlinks=true,
    linkcolor={red},
    citecolor={hanpurple},
    urlcolor={magenta}
}
\input{preamble}

\title{\LARGE \bf
Can we Trust Unreliable Voxels?\\Exploring 3D Semantic Occupancy Prediction under Label Noise
}  

\author{Wenxin Li$^{1}$, Kunyu Peng$^{2,3,*}$, Di Wen$^{2}$, Junwei Zheng$^{2}$, Jiale Wei$^{2}$, Mengfei Duan$^{1}$, Yuheng Zhang$^{1}$,\\Rui Fan$^{4}$, and Kailun Yang$^{1,*}$
\thanks{This work was supported in part by the National Natural Science Foundation of China (Grant No. 62473139), in part by the Hunan Provincial Research and Development Project (Grant No. 2025QK3019), in part by the State Key Laboratory of Autonomous Intelligent Unmanned Systems (the opening project number ZZKF2025-2-10), and in part by the Deutsche Forschungsgemeinschaft (DFG, German Research Foundation) - SFB 1574 - 471687386.}
\thanks{$^{1}$The authors are with the School of Artificial Intelligence and Robotics and the National Engineering Research Center of Robot Visual Perception and Control Technology, Hunan University, China (email: kailun.yang@hnu.edu.cn).}%
\thanks{$^{2}$The authors are with the Institute for Anthropomatics and Robotics, Karlsruhe Institute of Technology, Germany (email: kunyu.peng@kit.edu).}%
\thanks{$^{3}$The author is also with INSAIT, Sofia University ``St. Kliment Ohridski'', Bulgaria.}%
\thanks{$^{4}$The author is with the State Key Laboratory of Intelligent Autonomous Systems, Tongji University, Shanghai 201804, China.}
\thanks{*Corresponding authors: Kailun Yang and Kunyu Peng.}%
}

\let\oldtwocolumn\twocolumn
\renewcommand\twocolumn[1][]{%
    \oldtwocolumn[{#1}{
    \begin{center}
    \vskip-2ex
        \centering
        \includegraphics[width=0.99\textwidth]{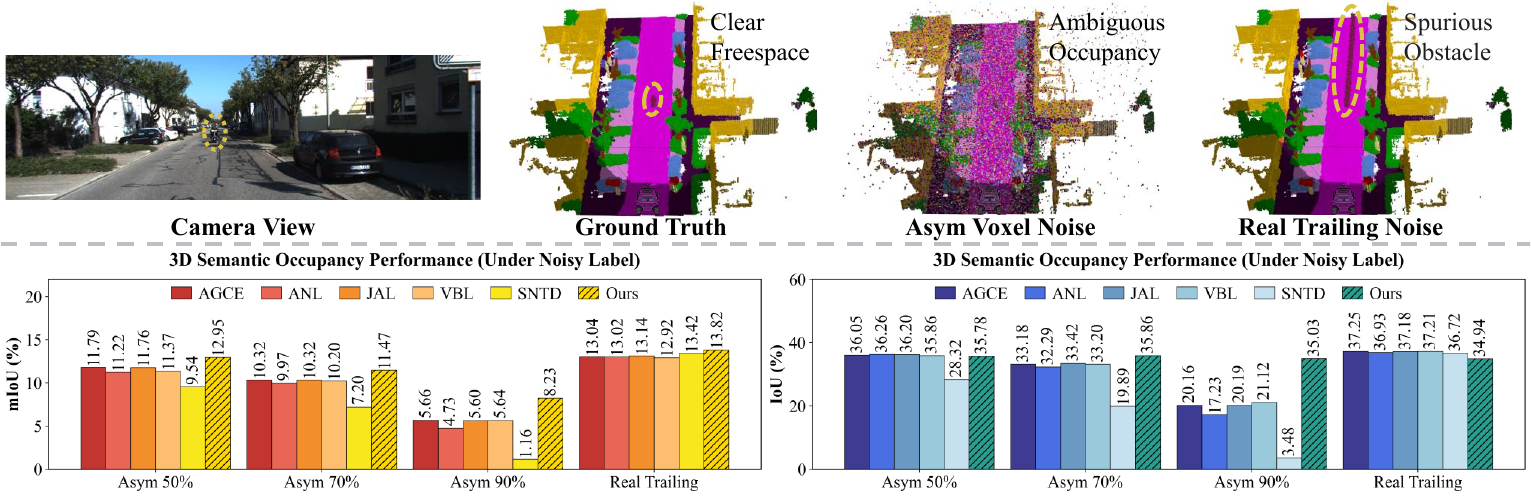}
        \vskip-1ex
        \captionof{figure}{{\textbf{Comparison of our proposed \textit{DPR-Occ} with state-of-the-art label-noise learning methods on our \textit{OccNL} benchmark.} 
        The upper part presents examples of semantic occupancy predictions under noisy supervision, including occupancy-asymmetric and real-world dynamic trailing noise. 
        The lower part shows that existing robust learning strategies struggle to alleviate the adverse effects of voxel-level noise, whereas our proposed noise-robust framework consistently improves semantic performance and achieves state-of-the-art mIoU under both synthetic and real-world noise settings, with particularly pronounced gains in extreme noise scenarios.
        }}
        \label{fig:teaser}
    \end{center}
    }]
}

\begin{document}

\maketitle
\thispagestyle{empty}
\pagestyle{empty}

\begin{abstract}
3D semantic occupancy prediction is a cornerstone of robotic perception, yet real-world voxel annotations are inherently corrupted by structural artifacts and dynamic trailing effects. This raises a critical but underexplored question: can autonomous systems safely rely on such unreliable occupancy supervision? To systematically investigate this issue, we establish \textit{OccNL}, the first benchmark dedicated to 3D occupancy under occupancy-asymmetric and dynamic trailing noise. Our analysis reveals a fundamental domain gap: state-of-the-art 2D label noise learning strategies collapse catastrophically in sparse 3D voxel spaces, exposing a critical vulnerability in existing paradigms. To address this challenge, we propose \textit{DPR-Occ}, a principled label-noise-robust framework that constructs reliable supervision through dual-source partial label reasoning. By synergizing temporal model memory with representation-level structural affinity, \textit{DPR-Occ} dynamically expands and prunes candidate label sets to preserve true semantics while suppressing noise propagation. Extensive experiments on SemanticKITTI demonstrate that \textit{DPR-Occ} prevents geometric and semantic collapse under extreme corruption. Notably, even at $90\%$ label noise, our method achieves significant performance gains (up to $2.57\%$ mIoU and $13.91\%$ IoU) over existing label noise learning baselines adapted to the 3D occupancy prediction task. By bridging label noise learning and 3D perception, \textit{OccNL} and \textit{DPR-Occ} provide a reliable foundation for safety-critical robotic perception in dynamic environments. The benchmark and source code will be made publicly available at \url{https://github.com/mylwx/OccNL}. 
\end{abstract}

\section{Introduction}
\input{Contents/Introduction}

\section{Related Work}

\input{Contents/Related_Work}
\section{OccNL: Established Benchmark}
\input{Contents/Benchmark}

\section{Methodology}
\input{Contents/Methodology}

\section{Experiments}
\label{sec:experiments}

\input{Contents/Experiments}

\section{Conclusion}
\input{Contents/Conclusion}

{\small
\bibliographystyle{IEEEtran}
\bibliography{bib}
}

\end{document}

%% file: preamble.tex
\definecolor{car}{rgb}{0.39215686, 0.58823529, 0.96078431}
\definecolor{bicycle}{rgb}{0.39215686, 0.90196078, 0.96078431}
\definecolor{motorcycle}{rgb}{0.11764706, 0.23529412, 0.58823529}
\definecolor{truck}{rgb}{0.31372549, 0.11764706, 0.70588235}
\definecolor{othervehicle}{rgb}{0.39215686, 0.31372549, 0.98039216}
\definecolor{person}{rgb}{1.        , 0.11764706, 0.11764706}
\definecolor{bicyclist}{rgb}{1.        , 0.15686275, 0.78431373}
\definecolor{motorcyclist}{rgb}{0.58823529, 0.11764706, 0.35294118}
\definecolor{road}{rgb}{1.        , 0.        , 1.        }
\definecolor{parking}{rgb}{1.        , 0.58823529, 1.        }
\definecolor{sidewalk}{rgb}{0.29411765, 0.        , 0.29411765}
\definecolor{otherground}{rgb}{0.68627451, 0.        , 0.29411765}
\definecolor{building}{rgb}{1.        , 0.78431373, 0.        }
\definecolor{fence}{rgb}{1.        , 0.47058824, 0.19607843}
\definecolor{vegetation}{rgb}{0.        , 0.68627451, 0.        }
\definecolor{trunk}{rgb}{0.52941176, 0.23529412, 0.        }
\definecolor{terrain}{rgb}{0.58823529, 0.94117647, 0.31372549}
\definecolor{pole}{rgb}{1.        , 0.94117647, 0.58823529}
\definecolor{trafficsign}{rgb}{1.        , 0.        , 0.        }


%% file: Contents/Introduction.tex
3D Semantic Occupancy Prediction~\cite{li2025voxdet}, also referred to as Semantic Scene Completion (SSC), aims to infer a dense voxel-grid representation of the environment by jointly predicting occupancy and semantic labels. 
As a foundational capability for autonomous driving~\cite{wang2025uniocc} and robotic perception~\cite{marcuzzi2025sfmocc,kim2025vpocc}, it serves as a fundamental building block, supporting downstream tasks including scene understanding~\cite{zuo2025quadricformer}, mapping~\cite{reed2025online}, and motion planning~\cite{shi2025h3o}. 
Recent studies~\cite{su2024alpha,chen2025semantic} have shown that voxelized 3D representations offer a practical paradigm for dense scene modeling, which has motivated a growing body of camera-based semantic occupancy prediction methods~\cite{li2025voxdet,marcuzzi2025sfmocc}. 
However, despite the rapid evolution of model architectures~\cite{kim2025protoocc}, the issue of voxel-wise label noise~\cite{li2025voxdet} remains largely overlooked.

In real-world applications, acquiring spatiotemporally aligned, perfectly consistent, and label noise-free 3D voxel annotations is a formidable challenge~\cite{zhou2025autoocc}.
Even within curated benchmark datasets, inherent flaws in the voxelization process, inaccuracies in cross-frame fusion, and the rapid motion of dynamic objects frequently introduce structural artifacts, such as trailing effects~\cite{behley2019semantickitti}. 
These imperfections lead to erroneous supervision during training and introduce significant evaluation bias during testing, as noted in VoxDet~\cite{li2025voxdet}. 
Such label uncertainty poses a fundamental threat to the safety and reliability of autonomous driving and robotic systems.
Based on this, a critical research question is raised: \textbf{Can we trust unreliable annotated voxels?} 

To answer this question, we present \textit{OccNL}, the first systematic benchmark exploring 3D semantic occupancy prediction under label noise, incorporating two types of noise: occupancy-asymmetric and real-world dynamic object trailing noise. 
Considering the heavy overhead of 3D occupancy, we adapt five state-of-the-art, memory-efficient label-noise-robust strategies from the image classification domain~\cite{zhou2023asymmetric,wang2025variation,ye2024active,wang2025joint,lan2025continuous}. 
However, extensive evaluations reveal their catastrophic degradation (\textit{e.g.}, all~\cite{zhou2023asymmetric,ye2024active,wang2025joint,wang2025variation,lan2025continuous} below $6\%$ mIoU and $22\%$ IoU at $90\%$ noise, see Fig.~\ref{fig:teaser}), failing to handle the inherent sparsity and irregularity of voxel data.

Motivated by the need for robustness against voxel-level label noise, we propose a label noise-robust 3D semantic occupancy prediction framework, Dual-source Partial-label Reasoning for Occupancy (\textit{DPR-Occ}). 
By synergizing temporal model-level memory and representation-level prototype affinity, \textit{DPR-Occ} dynamically constructs dual-source partial labels to mitigate error propagation via partial-label learning, negative learning, and EMA-guided self-distillation. 
Extensive evaluations demonstrate its consistent superiority: even under extreme $90\%$ label corruption where baselines collapse, \textit{DPR-Occ} preserves geometric and semantic integrity, yielding significant gains of up to $2.57\%$ mIoU and $13.91\%$ IoU over existing label noise learning baselines.

Our contributions are threefold:
\begin{itemize}
\item We introduce \textit{OccNL}, the first benchmark for 3D occupancy prediction under label noise. \textit{OccNL} incorporates both synthetic perturbations and real-world structured noise, and adapts multiple state-of-the-art label noise learning strategies from image-based visual recognition~\cite{zhou2023asymmetric,ye2024active,wang2025joint,wang2025variation,lan2025continuous} to 3D voxel paradigms.

\item We propose \textit{DPR-Occ}, a novel dual-source reasoning framework that dynamically fuses temporal model-level memory and prototype affinity to systematically mitigate voxel-level label corruption via partial-label learning, negative learning, and EMA-guided self-distillation.

\item Extensive experiments demonstrate that \textit{DPR-Occ} achieves state-of-the-art mIoU across synthetic and real-world noise scenarios on the \textit{OccNL} benchmark, successfully preserving geometric integrity and preventing the structural collapse that afflicts existing baseline methods under extreme noise. 

\end{itemize}

%% file: Contents/Related_Work.tex
\subsection{Semantic Occupancy Prediction}
Vision-based semantic occupancy prediction aims to infer the 3D occupancy and semantic status of space, typically represented as dense voxel grids~\cite{kim2025protoocc,li2025voxdet}. Beyond architectural designs, prior works focus on managing data unreliability through uncertainty quantification~\cite{heidrich2025occuq,wang2025reliable} or particle filtering~\cite{chen2025particle,deng2022_hd_ccsom}.
To mitigate annotation costs, other approaches leverage uncertainty-aware pseudo-labeling derived from sparse~\cite{liu2025sparse_annotation}, unlabeled~\cite{li2025enhancing_unlabeled}, or language-driven sources~\cite{yu2025language_driven,marcuzzi2025sfmocc}, utilizing confidence scores to filter noisy supervision.
However, a fundamental uncertainty remains within ground truth itself. While nuCraft~\cite{zhu2024nucraft} highlights severe noise in existing benchmarks and VoxDet~\cite{li2025voxdet} reveals that sequential afterimages in labels unfairly penalize reasonable predictions, a systematic investigation into occupancy label noise remains unexplored. 
To bridge this gap, we propose \textit{OccNL}, the first benchmark for 3D occupancy prediction under controllable corruptions.

\subsection{Noisy Label Learning}
Research on learning with noisy labels has flourished in the 2D image domain~\cite{zhou2023asymmetric,ye2024active,wang2025joint,wang2025variation,lan2025continuous}. Within this landscape, knowledge distillation has emerged as a powerful paradigm for noise robustness. 
Various strategies have been proposed, including two-stream knowledge distillation that refines labels through mutual agreement or sample sieving~\cite{jiang2023knowledge_distillation,bai2025robust_noisy_label_learning}, self-distillation mechanisms~\cite{das2023understanding_self_distillation}, and negative distillation strategies~\cite{lu2024federated}. 
Additionally, teacher-student architectures~\cite{tarvainen2017mean,gao2023semi,lan2025continuous} are widely adopted to provide stable supervision targets.
Unlike label noise in 2D images, 3D semantic occupancy prediction remains challenging due to severe sparsity and complex geometric artifacts~\cite{tang2024sparseocc}. 
This fundamental domain gap renders directly transferred 2D robust strategies prone to representation collapse. 
Within sparse 3D spaces, statistical methods struggle to delineate boundaries amidst voxel-level noise, frequently misidentifying minority semantic signals as outliers and hindering the learning of robust features. 
Since systematic studies on voxel-level label noise remain scarce, we introduce \textit{OccNL}, the first controllable noise benchmark analyzing the impacts of category flipping and dynamic object trailing in this field.

%% file: Contents/Benchmark.tex
\label{sec:benchmark}

\subsection{Task Definition}
\label{subsec:task_def}
\textit{OccNL} defines the task of 3D semantic \textbf{Occ}upancy prediction under \textbf{N}oisy \textbf{L}abels to evaluate label-noise robustness via a dual-protocol framework. 
We define the voxel label space $\mathcal{Y}{=}\{0,1,\dots,C\}$, where $0$ denotes the empty voxel and $\{1,\dots,C\}$ are semantic classes, totaling $C{+}1$ categories. The model learns a mapping $\mathcal{F}_\theta{:} X$→$\mathcal{Y}^{|\Omega|}$ over the voxel set $\Omega$. 
In \textit{Noisy Training}, the model is supervised by $\tilde{D}_{\text{train}}{=}\{(X_i, \tilde{Y}_i)\}_{i=1}^{N}$, where $\tilde{Y}_i$ denotes the noisy labels from occupancy-asymmetric flipping or real-world dynamic artifacts that may differ from the latent ground-truth $Y_i$. In \textit{Refined Evaluation}, performance is evaluated on a clean, refined set $D_{\text{eval}}{=}\{(X_j, Y^{(j)}_{\mathrm{refined}})\}_{j=1}^{M}$ to reduce bias.

\subsection{Ground Truth Refinement}

\label{subsec:gt_refinement}
Existing benchmarks like SemanticKITTI~\cite{behley2019semantickitti} suffer from dynamic object trailing artifacts due to multi-frame aggregation, leading to biased evaluation~\cite{li2025voxdet}. Following SCPNet~\cite{xia2023scpnet}, we construct a clean dataset by enforcing spatiotemporal consistency for dynamic categories $C_{dyn}$ (\textit{e.g.}, \textit{moving cars}) via instance masks $M_t$. A dynamic voxel $v$ is retained only if it aligns with the current object position:
\begin{equation}
    y^{\mathrm{refined}}_v = 
    \begin{cases} 
      0 & \text{if } y_v \in C_{dyn} \land v \notin M_t, \\
      y_v & \text{otherwise}.
    \end{cases}
\end{equation}
This refinement removes real-world trailing artifacts, ensuring strict evaluation against the current scene geometry.

\begin{figure}[tb]
  \centering
  \includegraphics[width=\linewidth]{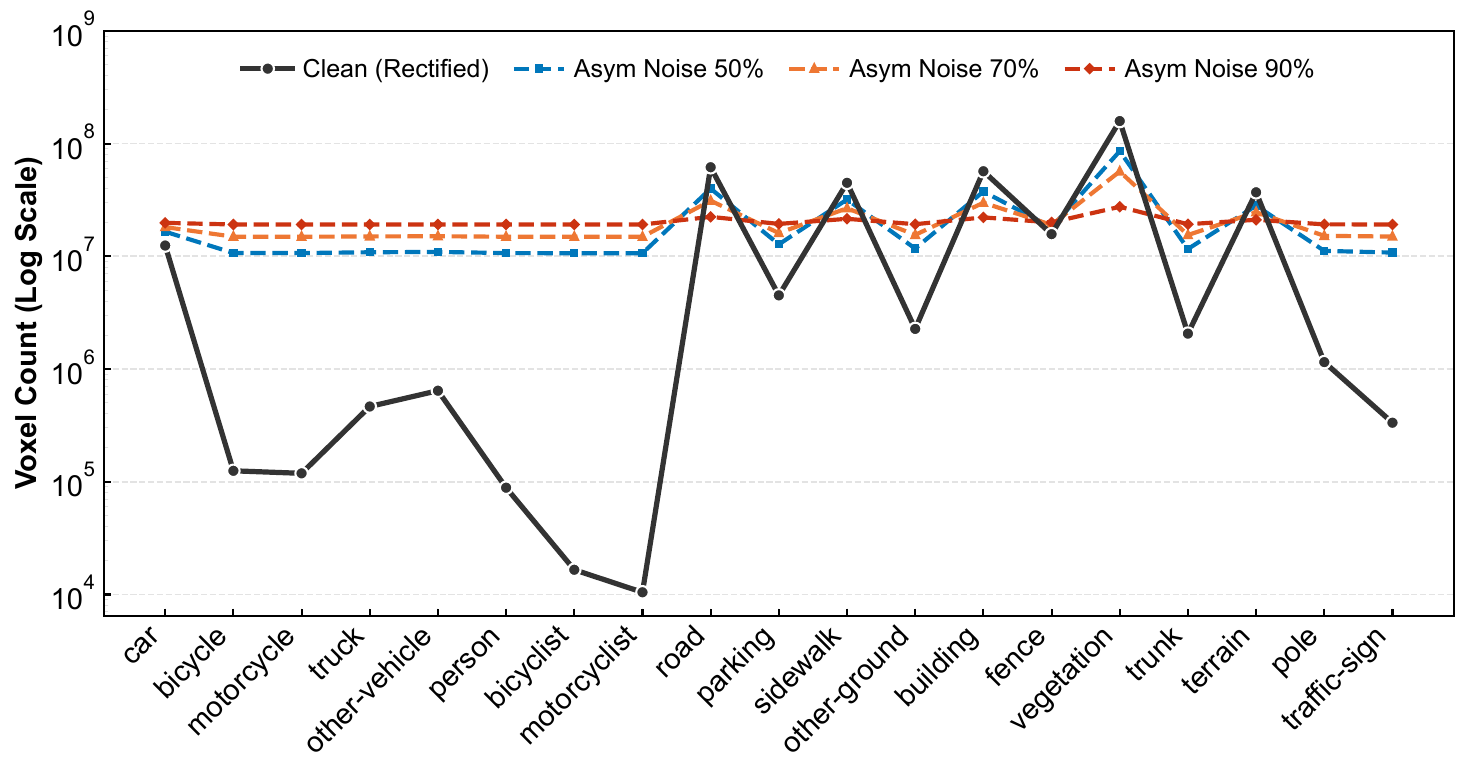}
  \vskip-1ex
  \caption{Semantic distribution evolution under voxel-level category-flipping noise. Due to orders-of-magnitude differences in voxel counts, we employ a logarithmic scale for visualization (empty voxels omitted as they remain constant at $10^{-3}\eta$ flip rate). Increased noise drives the distribution toward uniformity: dominant classes (\textit{e.g.}, \textit{vegetation}, \textit{road}) are suppressed, while rare classes (\textit{e.g.}, \textit{motorcyclist}, \textit{person}) are artificially inflated, culminating in nearly identical voxel frequencies at $90\%$ noise.}
  \label{fig:occ_asym_class_distribution}
  \vskip-3ex
\end{figure}

\input{table/protoocc_skitti_asym_noise_setting}
\subsection{Synthetic and Real-World Noise Setup}
\label{subsec:noise_synthesis}
\textit{Synthetic Occupancy-asymmetric Voxel Noise:}
To align with image-domain label noise research while preserving 3D geometric consistency, \textit{OccNL} introduces occupancy-asymmetric voxel noise. For each voxel $v {\in} \Omega$ (voxel set) with a noise rate $\eta$: 
Semantically occupied voxels ($y_v \in \mathcal{Y} \setminus \{0\}$) are flipped to another category (including empty) via $\tilde{y}_v {\sim} U(\mathcal{Y} {\setminus} \{y_v\})$, simulating category-flipping caused by semantic ambiguity in sparse, distant point clouds and in 2D-3D projection misalignments.  
To prevent structure collapse, empty voxels ($y_v{=}0$) are flipped to semantic categories with probability $10^{-3}\eta$, \textit{i.e.}, $\tilde{y}_v {\sim} U(\mathcal{Y} {\setminus} \{0\})$, reflecting spurious returns and clutter induced by adverse weather or multi-path effects. 
Empirically, while geometric completeness (IoU) remains robust at $50\%$ noise, semantic discrimination (mIoU) degrades significantly compared to the $20\%$ setting (Table~\ref{tab:protoocc_skitti_asym_noise_setting}). 
Thus, we set $\eta {\in} \{0.5, 0.7, 0.9\}$ as light, moderate, and heavy noise. Notably, the $90\%$ setting defines an extremely challenging scenario that exceeds typical sensor errors and long-range sparsity (Fig.~\ref{fig:occ_asym_class_distribution}), acting as a stress test for robust label-noise learning strategies under catastrophic corruption.

\textit{Real-World Dynamic Object Trailing Noise:}
Targeting spatiotemporal inconsistency from dynamic objects ($<5.21\%$ of voxels) in SemanticKITTI~\cite{behley2019semantickitti}, we construct three trailing noise levels. Empirically, the future $70$ frames cover most motion trajectories within the field of view, while $70$ superimposed historical frames span the perceptual temporal domain. Based on this, we define: \textit{Mild}: uncompensated fused data from future $70$ frames, consistent with standard SemanticKITTI~\cite{behley2019semantickitti}; \textit{Moderate}: integration of $70$ historical frames for an enhanced full-domain trailing effect; \textit{Severe}: additional voxel dilation ($6$-connectivity) on dynamic voxels to simulate extreme high-density trailing corruption.

%% file: table/protoocc_skitti_asym_noise_setting.tex
\begin{table*}[!t]
\newcommand{\clsname}[2]{
  \rotatebox{90}{
    \hspace{-6pt}
    \textcolor{#2}{$\blacksquare$}
    \hspace{-6pt}
    \begin{tabular}{l}
      #1                                      
    \end{tabular}
  }
}
\centering
\caption{Quantitative evaluation of ProtoOcc~\cite{kim2025protoocc} on the OccNL benchmark under different levels of asymmetric voxel noise. 
The model is trained with varying noise rates $\eta \in \{0\%, 20\%, 50\%, 70\%, 90\%\}$ to simulate semantic annotation interference. Note that while geometric IoU remains relatively stable at low noise levels, a significant degradation in mIoU and fine-grained category performance is observed as $\eta$ reaches $50\%$ and beyond, reflecting the backbone's threshold for interference suppression.}
\label{tab:protoocc_skitti_asym_noise_setting}
\vskip-1ex
\small
\resizebox{\linewidth}{!}{
\begin{tabular}{l|cc|ccccccccccccccccccc}
\toprule
\makecell{Noise \\ Rate} & 
{\rotatebox{90}{IoU}} & 
{\rotatebox{90}{mIoU}} & 
\clsname{road}{road} &
\clsname{sidewalk}{sidewalk} &
\clsname{parking}{parking} &
\clsname{other-grnd.}{otherground} &
\clsname{building}{building} &
\clsname{car}{car} &
\clsname{truck}{truck} &
\clsname{bicycle}{bicycle} &
\clsname{motorcycle}{motorcycle} &
\clsname{other-veh.}{othervehicle} &
\clsname{vegetation}{vegetation} &
\clsname{trunk}{trunk} &
\clsname{terrain}{terrain} &
\clsname{person}{person} &
\clsname{bicyclist}{bicyclist} &
\clsname{motorcyclist}{motorcyclist} &
\clsname{fence}{fence} &
\clsname{pole}{pole} &
\clsname{traf.-sign}{trafficsign} \\
\midrule
$0\%$  & 37.39 & 13.85 & 59.76 & 28.29 & 22.28 & 0.00 & 15.15 & 28.48 & 5.19 & 2.29 & 2.99 & 11.30 & 19.11 & 3.44 & 31.73 & 5.93 & 13.23 & 0.00 & 6.53 & 4.48 & 2.89 \\
$20\%$ & 36.62 & 14.20 & 59.05 & 28.72 & 21.76 & 0.00 & 14.95 & 29.76 & 21.89 & 0.41 & 3.82 & 13.77 & 18.39 & 2.63 & 32.05 & 4.92 & 9.24 & 0.00 & 7.05 & 1.42 & 0.00 \\
$50\%$ & 36.00 & 12.52 & 58.42 & 28.41 & 20.94 & 0.42 & 14.52 & 28.91 & 16.72 & 0.00 & 0.00 & 12.31 & 17.95 & 1.45 & 30.62 & 0.00 & 0.00 & 0.00 & 7.19 & 0.00 & 0.00 \\
$70\%$ & 33.39 & 10.73 & 56.56 & 28.92 & 18.65 & 0.00 & 13.69 & 24.95 & 10.82 & 0.00 & 0.00 & 0.00 & 15.07 & 0.00 & 29.59 & 0.00 & 0.00 & 0.00 & 5.67 & 0.00 & 0.00 \\
$90\%$ & 21.20 & 6.41  & 47.70 & 23.13 & 5.02  & 0.00 & 5.56  & 13.59 & 0.00 & 0.00 & 0.00 & 0.00 & 5.92  & 0.00 & 19.78 & 0.00 & 0.00 & 0.00 & 1.02 & 0.00 & 0.00 \\
\bottomrule
\end{tabular}
}
\vskip-3ex
\end{table*}

%% file: Contents/Methodology.tex
\begin{figure*}[!t]
  \centering%
  \includegraphics[width=\textwidth]{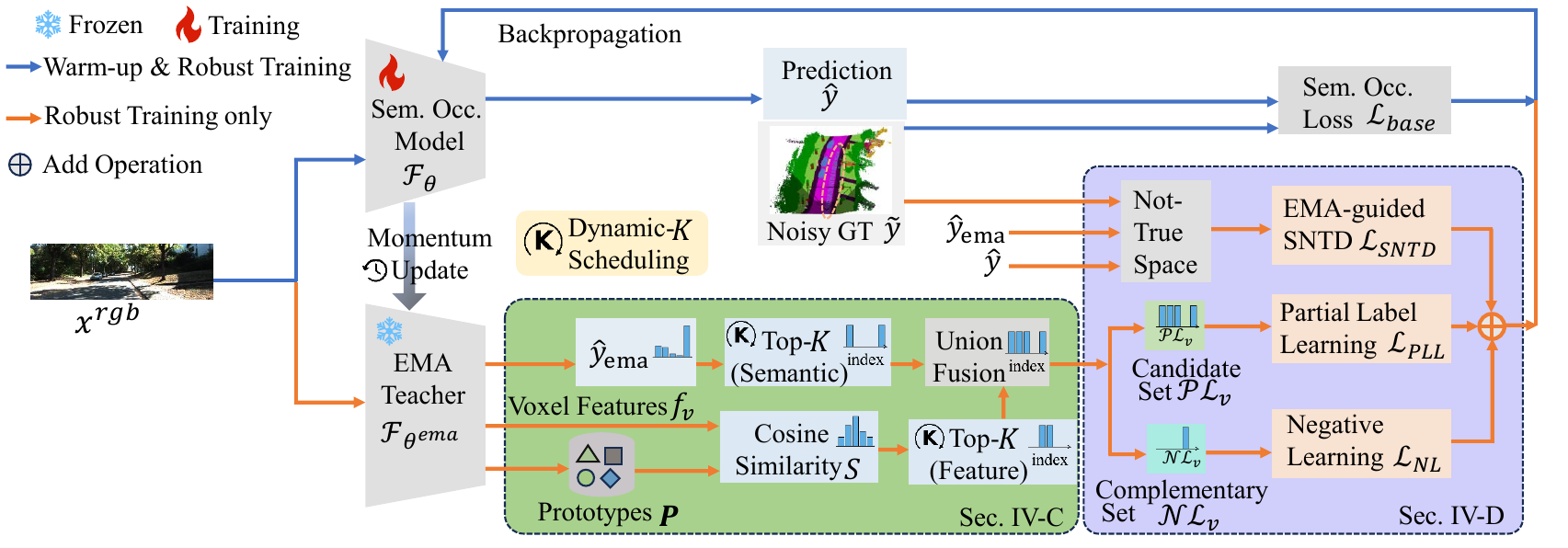}
  \vskip-1ex
\caption{The overall framework of our proposed \textit{DPR-Occ}. \textit{Warm-up Stage}: The model captures clean patterns via standard training on noisy labels $\tilde{Y}$. \textit{Robust Stage}: Guided by dynamic-$K$ scheduling, we construct dual-source partial label sets by fusing Top-$K$ predictions from the EMA teacher and feature-prototype similarities. The network is then optimized using Partial Label Learning (PLL) and Negative Learning (NL), with EMA-guided Self-Not-True Distillation (SNTD) further suppressing noise for robust learning.
}
\label{fig:occnl_framework}
\vskip-3ex
\end{figure*}

\subsection{Overview of the Proposed Framework}
\label{subsec:framework_overview}
As illustrated in Fig.~\ref{fig:occnl_framework}, \textit{DPR-Occ} employs a two-stage strategy to optimize the 3D semantic occupancy model $\mathcal{F}_\theta$.
\textit{Warm-up Stage}: The model learns clean patterns via the memorization effect~\cite{arpit2017closer} under noisy supervision, while synchronously updating the EMA teacher. 
\textit{Robust Learning Stage}: With dual-source partial labels fused from Top-$K$ EMA predictions and prototype similarities, $\mathcal{F}_\theta$ is optimized via Partial Label Learning (PLL) and Negative Learning (NL), with Self-Not-True Distillation (SNTD) regularizing non-target classes to suppress noise, as detailed in Alg.~\ref{alg:occnl}.

\subsection{Warm-Up Stage}
The model is trained directly on the corrupted dataset $\tilde{D}_{\text{train}}$ (Sec.~\ref{subsec:noise_synthesis}) by optimizing the standard supervised loss $\mathcal{L}_{base}$ (\textit{e.g.}, ProtoOcc~\cite{kim2025protoocc}) without robust strategies, exploiting the DNN memorization effect to learn clean patterns~\cite{arpit2017closer}. Simultaneously, the EMA teacher $\theta^{ema}$ is updated at each iteration $t$ via momentum:
\begin{equation}
    \theta_{t}^{ema} = d_t \theta_{t-1}^{ema} + (1 - d_t) \theta_{t}.
    \label{eq:ema_update}
\end{equation}
Consistent with ProtoOcc~\cite{kim2025protoocc}, we maintain class prototypes~$P{=}\{\boldsymbol{p}_c\}_{c=0}^C$, where each $\boldsymbol{p}_c$ fuses a scene-adaptive prototype (via class-wise average pooling over current-scene voxel features) with a scene-agnostic prototype (accumulated from scene-adaptive prototypes across training samples), thereby capturing global semantic distributions to facilitate subsequent partial label construction.

\subsection{Dual-Source Partial Label Construction}
To enhance the ground-truth hit rate while constraining candidate space expansion under noisy annotations, we integrate semantic evidence from EMA teacher consensus and representation-structure evidence via feature-prototype similarity. This dual-source construction expands the candidate set to encompass true labels while constraining the disambiguation space, thereby mitigating error propagation and accumulation triggered by label noise.

\textit{EMA Teacher Consensus as Semantic Evidence:}
For voxel $v {\in} \Omega$, the EMA teacher network outputs the category probability distribution $\boldsymbol{y}_v^{ema}$. Compared to the noisy supervision $\tilde{y}_v$, the EMA teacher provides a stable and noise-resistant semantic consensus. 
By integrating historical prediction information at the model level, it effectively mitigates prediction oscillations induced by voxel-level label noise.

\textit{Prototype Affinity as Representation-Structure Evidence:}
To complement semantic consensus, we enforce structural consistency by measuring the cosine similarity between voxel feature $\boldsymbol{f}_v$ and category prototype $\boldsymbol{p}_c\!\in\!P$, both extracted from the EMA teacher:
\begin{equation}
S(v, c) = \frac{\boldsymbol{f}_v \cdot \boldsymbol{p}_c}{\|\boldsymbol{f}_v\| \|\boldsymbol{p}_c\|}.
\label{eq:similarity}
\end{equation}
The structural candidate set is formed by selecting the top-$K$ categories with the highest similarity, characterizing the feature-prototype consistency.

\textit{Union Fusion for Candidate Set Construction:}
We construct the partial label candidate set $\mathcal{PL}_v$ by taking the union of Top-$K$ categories from both EMA teacher predictions and prototype similarity scores:
\begin{equation}
\mathcal{PL}_v = \operatorname{TopK}(\boldsymbol{y}_v^{ema}, K_e) \cup \operatorname{TopK}(S(v,\cdot), K_e).
\label{eq:union_fusion}
\end{equation}
This union mechanism leverages semantic-structural complementarity to retain the ground truth even if its rank drops in one source, thereby improving coverage and reducing error accumulation during disambiguation.

\subsection{Dynamic-$K$ Scheduling and Joint Training Objective}
In noisy learning, the candidate set size $K$ governs the trade-off between ground-truth hit rate and candidate purity. We employ a dynamic $K$ strategy: using a larger $K$ early in training to maximize coverage, and gradually decreasing $K$ to minimize disambiguation difficulty and enhance purity. A robust joint optimization objective is then constructed based on this scheduling.

Let $e$ be the current epoch and $E_{w}$ the warm-up duration. After the warm-up stage, the candidate set size $K_e$ is linearly decayed to balance hit rate and purity:
\begin{equation}
K_e = \max \left( K_{end},  K_{start} - \gamma \cdot (e - E_{w} - 1) \right),
\label{eq:dynamic_k}
\end{equation}
where $K_{start}, K_{end}$ are the initial and terminal sizes, and $\gamma$ is the step size. This strategy ensures early-stage ground-truth encompassment via a larger $K_e$, while narrowing the search space as model certainty increases to alleviate disambiguation complexity.

\textit{Partial Label Learning Loss:}
To guide the student $\mathcal{F}_\theta$ in allocating probability mass within the candidate set $\mathcal{PL}_v$, we optimize the PLL loss $\mathcal{L}_{PLL}$:
\begin{equation}
\mathcal{L}_{PLL} = - \frac{1}{|\Omega|} \sum_{v \in \Omega} \frac{1}{|\mathcal{PL}_v|} \sum_{c \in \mathcal{PL}_v} \log p_{stu, v}(c),
\label{eq:loss_pll}
\end{equation}
where $p_{stu, v}(c)$ denotes the softmax probability of category $c$ at voxel $v$ predicted by the student model.

\textit{Negative Learning on Complementary Set:}
To suppress noise outside the candidate set, we define the complementary set $\mathcal{NL}_v{=}\{c | c\notin\mathcal{PL}_v\}$ and optimize the NL loss $\mathcal{L}_{NL}$:
\begin{equation}
\mathcal{L}_{NL} = - \frac{1}{|\Omega|} \sum_{v \in \Omega} \frac{1}{|\mathcal{NL}_v|} \sum_{c \in \mathcal{NL}_v} \log (1 - p_{stu, v}(c)),
\label{eq:loss_nl}
\end{equation}
which penalizes categories deemed unreliable by both the EMA teacher and structural prototypes.

\textit{Distribution Alignment Regularization in Not-True Space:}
To further shield the model from noisy supervision $\tilde{y}_v$, we adapt Self-Not-True Distillation (SNTD)~\cite{lan2025continuous} as a regularizer. Unlike~\cite{lan2025continuous}, we leverage the EMA teacher to provide stable model-level historical supervision. We compute the not-true distribution $\tilde{\boldsymbol{p}}_{v}$ by masking the noisy label $\tilde{y}_v$ and re-normalizing the remaining $C$ logits:
\begin{equation}
\tilde{p}_{v,k} = \frac{\exp(z_{v,k} / \tau_s)}{\sum_{j \neq \tilde{y}_v} \exp(z_{v,j} / \tau_s)}, \quad \forall k \neq \tilde{y}_v,
\label{eq:not_true_dist}
\end{equation}
where $z_{v,k}$ is the logit and $\tau_s$ denotes the temperature. $\mathcal{L}_{SNTD}$ is the KL-divergence between the student and EMA teacher in this filtered space, both following Eq.~\eqref{eq:not_true_dist}:
\begin{equation}
\mathcal{L}_{SNTD} = \tau_s^{2} \sum_{v \in \Omega} D_{\mathrm{KL}}(\tilde{\boldsymbol{p}}_{v}^{ema} \parallel \tilde{\boldsymbol{p}}_{v}^{stu}).
\label{eq:loss_sntd}
\end{equation}
This strategy prevents the student from overfitting to the semantics of noisy labels while preserving global consistency from the teacher.

\input{table/protoocc_skitti_asym_val}

\subsection{Overall Training Objective}
\label{subsec:joint_objective}
To unify our label-noise robust strategies with 3D occupancy prediction, we define $\mathcal{L}_{robust}$ as follows:
\begin{equation}
    \mathcal{L}_{robust} = \mathcal{L}_{PLL} + \mathcal{L}_{NL} + \mathcal{L}_{SNTD}.
    \label{eq:loss_robust}
\end{equation}
The total objective $\mathcal{L}_{total}$ integrates the loss $\mathcal{L}_{base}$ with the robust module activated via the indicator function $\mathbb{I}(\cdot)$:
\begin{equation}
    \mathcal{L}_{total} = \mathcal{L}_{base} + \mathbb{I}(e > E_w) \cdot \mathcal{L}_{robust}.
    \label{eq:loss_total}
\end{equation}
This joint objective ensures that the model learns foundational geometry from $\mathcal{L}_{base}$ while $\mathcal{L}_{robust}$ prevents overfitting to corrupted labels (see Alg.~\ref{alg:occnl}), leading to the superior performance reported in Tables~\ref{tab:protoocc_skitti_asym_val} and~\ref{tab:protoocc_skitti_tail_val}.

%% file: table/protoocc_skitti_asym_val.tex
\begin{table*}[t]
\newcommand{\clsname}[2]{%
  \rotatebox{90}{%
    \hspace{-6pt}\textcolor{#2}{$\blacksquare$}\hspace{-6pt}%
    \begin{tabular}{l}#1\end{tabular}%
  }%
}

\centering
\caption{Comparison on \textit{OccNL} benchmark under occupancy-asymmetric voxel noise. We compare \textit{DPR-Occ} against image-domain robust strategies including AGCE~\cite{zhou2023asymmetric}, ANL~\cite{ye2024active}, JAL~\cite{wang2025joint}, VBL~\cite{wang2025variation}, and SNTD~\cite{lan2025continuous}. \textit{DPR-Occ} consistently maintains superior mIoU, especially under extreme $90\%$ noise.}
\label{tab:protoocc_skitti_asym_val}
\vskip-1ex
\small
\resizebox{\linewidth}{!}{
\begin{tabular}{c|l|cc|ccccccccccccccccccc}
\toprule
\makecell{Noise \\ Rate} &
Method &
{\rotatebox{90}{IoU}} &
{\rotatebox{90}{mIoU}} &
\clsname{road}{road} &
\clsname{sidewalk}{sidewalk} &
\clsname{parking}{parking} &
\clsname{other-grnd.}{otherground} &
\clsname{building}{building} &
\clsname{car}{car} &
\clsname{truck}{truck} &
\clsname{bicycle}{bicycle} &
\clsname{motorcycle}{motorcycle} &
\clsname{other-veh.}{othervehicle} &
\clsname{vegetation}{vegetation} &
\clsname{trunk}{trunk} &
\clsname{terrain}{terrain} &
\clsname{person}{person} &
\clsname{bicyclist}{bicyclist} &
\clsname{motorcyclist}{motorcyclist} &
\clsname{fence}{fence} &
\clsname{pole}{pole} &
\clsname{traf.-sign}{trafficsign} \\
\midrule

\multirow{6}{*}{$50\%$}
& AGCE & 36.05 & 11.79 & 58.66 & 29.08 & 19.32 & 0.00 & 15.22 & 28.17 & 6.53 & 0.00 & 0.00 & 12.28 & 17.46 & 0.00 & 29.49 & 0.00 & 0.00 & 0.00 & 7.70 & 0.00 & 0.00 \\
& ANL  & 36.26 & 11.22 & 58.03 & 29.90 & 17.25 & 0.00 & 16.58 & 26.31 & 0.91 & 0.00 & 0.00 & 7.88 & 17.67 & 0.00 & 31.16 & 0.00 & 0.00 & 0.00 & 7.42 & 0.00 & 0.00 \\
& JAL & 36.20 & 11.76 & 58.34 & 29.31 & 19.46 & 0.00 & 16.10 & 27.92 & 7.37 & 0.00 & 0.00 & 8.97 & 17.50 & 0.00 & 31.03 & 0.00 & 0.00 & 0.00 & 7.48 & 0.00 & 0.00 \\
& VBL & 35.86 & 11.37 & 57.72 & 28.47 & 19.73 & 0.19 & 16.00 & 27.70 & 1.42 & 0.00 & 0.00 & 10.66 & 16.83 & 0.00 & 30.62 & 0.00 & 0.00 & 0.00 & 6.74 & 0.00 & 0.00 \\
& SNTD & 28.32 & 9.54 & 58.06 & 30.16 & 15.75 & 0.00 & 9.64 & 28.15 & 0.03 & 0.00 & 0.00 & 0.00 & 9.90 & 0.00 & 27.19 & 0.00 & 0.00 & 0.00 & 2.29 & 0.00 & 0.00 \\
\rowcolor[gray]{.9} \multicolumn{1}{c|}{\cellcolor{white}} & Ours & 35.78 & 12.95 & 58.27 & 29.35 & 20.55 & 0.95 & 15.00 & 28.31 & 16.33 & 0.17 & 2.51 & 7.92 & 19.26 & 1.90 & 31.99 & 0.11 & 6.04 & 0.00 & 7.46 & 0.01 & 0.00 \\
\midrule

\multirow{6}{*}{$70\%$}
& AGCE & 33.18 & 10.32 & 56.94 & 28.95 & 19.44 & 0.00 & 13.41 & 23.16 & 2.68 & 0.00 & 0.00 & 0.03 & 14.73 & 0.00 & 30.17 & 0.00 & 0.00 & 0.00 & 6.54 & 0.00 & 0.00 \\
& ANL & 32.29 & 9.97 & 56.03 & 28.63 & 17.60 & 0.00 & 12.48 & 24.06 & 0.01 & 0.00 & 0.00 & 0.42 & 14.39 & 0.00 & 29.60 & 0.00 & 0.00 & 0.00 & 6.24 & 0.00 & 0.00 \\
& JAL & 33.42 & 10.32 & 57.84 & 28.98 & 18.71 & 0.00 & 13.89 & 24.43 & 0.00 & 0.00 & 0.00 & 0.01 & 14.84 & 0.00 & 30.78 & 0.00 & 0.00 & 0.00 & 6.53 & 0.00 & 0.00 \\
& VBL & 33.20 & 10.20 & 57.27 & 28.74 & 18.00 & 0.00 & 13.49 & 24.74 & 0.34 & 0.00 & 0.00 & 0.28 & 14.06 & 0.00 & 29.87 & 0.00 & 0.00 & 0.00 & 7.03 & 0.00 & 0.00 \\
& SNTD & 19.89 & 7.20 & 51.97 & 18.90 & 20.14 & 0.00 & 0.03 & 12.52 & 0.00 & 0.00 & 0.00 & 0.00 & 7.61 & 0.00 & 24.65 & 0.00 & 0.00 & 0.00 & 0.92 & 0.00 & 0.00 \\
\rowcolor[gray]{.9} \multicolumn{1}{c|}{\cellcolor{white}} & Ours & 35.86 & 11.47 & 57.97 & 29.00 & 19.30 & 0.40 & 14.94 & 28.13 & 10.66 & 0.01 & 0.00 & 0.85 & 18.98 & 0.00 & 31.46 & 0.00 & 0.00 & 0.00 & 6.27 & 0.00 & 0.00 \\
\midrule

\multirow{6}{*}{$90\%$}
& AGCE & 20.16 & 5.66 & 44.87 & 20.09 & 0.00 & 0.00 & 5.97 & 8.96 & 0.00 & 0.00 & 0.00 & 0.00 & 6.24 & 0.00 & 19.38 & 0.00 & 0.00 & 0.00 & 2.02 & 0.00 & 0.00 \\
& ANL & 17.23 & 4.73 & 41.58 & 18.86 & 0.00 & 0.00 & 2.26 & 7.72 & 0.00 & 0.03 & 0.00 & 0.00 & 4.24 & 0.00 & 14.81 & 0.00 & 0.00 & 0.00 & 0.45 & 0.00 & 0.00 \\
& JAL & 20.19 & 5.60 & 45.72 & 20.29 & 0.00 & 0.00 & 4.91 & 8.30 & 0.00 & 0.00 & 0.00 & 0.00 & 6.23 & 0.01 & 18.31 & 0.00 & 0.00 & 0.00 & 2.59 & 0.00 & 0.00 \\
& VBL & 21.12 & 5.64 & 43.75 & 19.65 & 0.00 & 0.00 & 5.86 & 10.38 & 0.00 & 0.00 & 0.00 & 0.03 & 7.74 & 0.00 & 17.77 & 0.00 & 0.00 & 0.01 & 2.03 & 0.00 & 0.00 \\
& SNTD & 3.48 & 1.16 & 21.86 & 0.01 & 0.01 & 0.00 & 0.00 & 0.09 & 0.00 & 0.00 & 0.00 & 0.00 & 0.00 & 0.00 & 0.00 & 0.00 & 0.00 & 0.00 & 0.00 & 0.00 & 0.00 \\
\rowcolor[gray]{.9} \multicolumn{1}{c|}{\cellcolor{white}} & Ours & 35.03 & 8.23 & 52.14 & 26.70 & 0.20 & 0.00 & 14.62 & 14.78 & 0.00 & 0.00 & 0.00 & 0.00 & 16.73 & 0.00 & 29.96 & 0.02 & 0.00 & 0.00 & 1.30 & 0.01 & 0.00 \\
\bottomrule
\end{tabular}
}
\vskip-3ex
\end{table*}

%% file: Contents/Experiments.tex
\subsection{Datasets and Metrics}
\textit{Datasets.} 
Following the unified setup~\cite{kim2025protoocc,li2025voxdet}, we conduct experiments on the SemanticKITTI dataset~\cite{behley2019semantickitti}, restricting
\input{Contents/alg/alg_occnl}
\noindent the study to the standard spatial volume of $[0, 51.2] \times [-25.6, 25.6] \times [-2.0, 4.4] \text{ m}^3$, voxelized into a $256 \times 256 \times 32$ grid ($0.2$\,m resolution) with $C{=}19$ semantic categories and one empty category (\textit{i.e.}, $C{+}1{=}20$ total categories). 
For data partitioning, we adopt the \textit{OccNL} benchmark settings, utilizing sequences 00--07 and 09--10 ($3{,}834$ frames) for training and sequence 08 ($815$ frames) for validation.
Details regarding label refinement and noise setup are provided in Sec.~\ref{subsec:gt_refinement} and Sec.~\ref{subsec:noise_synthesis}, respectively.

\input{table/protoocc_skitti_tail_val}
\textit{Metrics.} 
Following~\cite{xia2023scpnet,li2025voxdet,kim2025protoocc}, we evaluate performance using the Intersection over Union (IoU) for geometric completion and the mean IoU (mIoU) for semantic completion. The geometric IoU is calculated by treating all occupied voxels as one class. For semantic completion, the IoU for a class $i$ is defined as $IoU_i = \frac{TP_i}{TP_i + FP_i + FN_i}$, where $TP_i$, $FP_i$, and $FN_i$ denote the true positives, false positives, and false negatives for class $i$, respectively. Then, mIoU is computed by averaging $IoU_i$ across all semantic classes, serving as the primary metric for 3D semantic occupancy prediction.

\subsection{Implementation Details}
We employ ResNet-50~\cite{he2016deep} as the image backbone and train on $4 \times$ RTX 3090 for $20$ epochs (batch size $16$). 
The learning rate starts at $4{\times}10^{-4}$, decaying by $0.1$ at the $18$th epoch. 
The warm-up duration $E_{w}$ is $12$ epochs, and the EMA momentum $d_t$ is $0.999$. 
$K_e$ follows a linear decay with $K_{start}{=}9$, $K_{end}{=}2$, and step-size $\gamma{=}2$. The temperature $\tau_s$ in $\mathcal{L}_{SNTD}$ is $3.0$. 
The other configurations are consistent with ProtoOcc~\cite{kim2025protoocc} to ensure a fair comparison of both performance and inference efficiency.

\subsection{Baselines}
\label{subsec:baselines}

To pioneer the study of noisy labels in 3D semantic occupancy prediction, we establish the \textit{OccNL} benchmark. For a fair and wide comparison, we adapt five state-of-the-art 2D robust methods (AGCE~\cite{zhou2023asymmetric}, ANL~\cite{ye2024active}, JAL~\cite{wang2025joint}, VBL~\cite{wang2025variation}, SNTD~\cite{lan2025continuous}) to the 3D voxel domain and explore their optimal data-sensitive configurations.
Specifically, we replace standard cross-entropy objectives with each respective robust label noise learning loss across all voxel-wise and query-driven semantic prediction modules~\cite{kim2025protoocc}.

\subsection{Influence of Occupancy-Asymmetric Voxel Label Noise}
\label{subsec:influence_asym}
Tables~\ref{tab:protoocc_skitti_asym_noise_setting} and~\ref{tab:protoocc_skitti_asym_val} reveal a two-stage degradation regime in representation stability under asymmetric voxel noise.
In the early regime ($\eta \leq 50\%$), the baseline geometric IoU remains largely intact, while semantic discrimination for low-occupancy categories (\textit{e.g.}, \textit{bicycle}, \textit{pole}) collapses. 
This indicates that structural occupancy cues are more resilient than category-level semantics.
However, beyond a critical threshold ($\eta \geq 70\%$), geometry also deteriorates, transitioning from semantic confusion to structural instability.

This catastrophic degradation highlights a fundamental domain gap. As shown in Table~\ref{tab:protoocc_skitti_asym_val}, image-domain label-noise learning strategies exhibit near-extinction on the sparsest dynamic categories (\textit{e.g.}, \textit{bicyclist}, \textit{motorcyclist}) under extreme corruption ($90\%$). 
Methods like AGCE and JAL fail because high-rate voxel flipping actively reshapes the semantic distribution toward artificial uniformity, erasing the natural long-tailed structure of real-world environments. 
This is consistent with the distribution drift illustrated in Fig.~\ref{fig:occ_asym_class_distribution}, where voxel frequencies are pushed toward uniformity under high corruption. 
Their loss-level reweighting mechanisms implicitly amplify empty-voxel dominance, leading to a total collapse of sparse dynamic categories. 
Similarly, VBL's gradient variance constraint proves ineffective, as 3D data sparsity naturally induces high variance in minority classes. 
Furthermore, correction-based SNTD collapses under $90\%$ noise, as it relies on the model from the previous epoch for correction, which lacks the temporal stability of an Exponential Moving Average (EMA) teacher and fails to provide reliable supervision under severe interference.

In contrast, \textit{DPR-Occ} effectively raises the collapse threshold, maintaining a high geometric IoU of $35.03\%$ even at $90\%$ noise, demonstrating a stabilizing prior that preserves the scene's structural integrity and sparse semantics against baseline collapse and minority extinction.

\subsection{Effects of Dynamic Object Trailing Noise}
\label{subsec:influence_tail}

Dynamic object trailing noise (Table~\ref{tab:protoocc_skitti_tail_val}) simulates spatiotemporal inconsistencies commonly found in aggregated point cloud sequences.
Unlike asymmetric flipping, it introduces spatially coherent but semantically inconsistent voxel clusters, forming``ghost geometry'' that challenges models to distinguish between structural persistence and motion-induced artifacts. As trailing severity escalates from Mild to Severe, most baselines degrade visibly.
For instance, AGCE's mIoU drops ($13.04\%$→$11.58\%$), and SNTD's \textit{truck} performance plummets ($16.07\%$→$4.38\%$), indicating that existing methods struggle to separate true moving objects from ``ghost'' voxels.
Conversely, \textit{DPR-Occ} remains stable with the highest mIoU across all settings with negligible fluctuation ($13.82\%$→$13.67\%$), confirming its ability to effectively disentangle semantic features from structural noise.

Trailing artifacts disproportionately corrupt motion-sensitive cues in small and sparse classes, \textit{e.g.}, \textit{person}, \textit{bicyclist}. 
Under the \textit{Severe} setting, several baselines yield very low IoU on \textit{bicyclist} and exhibit limited performance on \textit{person}.
However, \textit{DPR-Occ} retains robust discriminability, achieving $3.95\%$ IoU for \textit{bicyclist} and maintaining competitive performance on major dynamic categories, \textit{e.g.}, \textit{car}, \textit{truck}.
These results confirm that our strategy learns to extract the intrinsic geometry of dynamic targets rather than merely overfitting to the noise distribution.

\input{table/ablation_Ew}
\subsection{Ablation Studies}

\textit{Ablation Study of Warm-up Duration.}
Evaluated at $70\%$ noise (Table~\ref{tab:ablation_ew}), while mIoU remains relatively stable between $6$ and $16$ epochs, the IoU exhibits a significant improvement when increasing $E_{w}$ from $6$ to $12$ epochs. 
Further extending $E_{w}$ to $14$ epochs yields only marginal gains in IoU but leads to a noticeable drop in mIoU. 
This suggests that $12$ epochs provide the optimal trade-off for representation learning under label corruption. 

\textit{Ablation of Dynamic-$K$ Scheduling.} 
We compare Linear scheduling (see Eq. (\ref{eq:dynamic_k})) against Fixed and Random baselines (Table~\ref{tab:ablation_dynamic_k}). 
Under $70\%$ noise, a large set (Fixed $9$) yields the best mIoU ($11.56\%$),
as improved ground-truth coverage outweighs the loss in disambiguation.
However, at $50\%$ noise, disambiguation becomes critical, causing Fixed $9$ to degrade to $11.48\%$. Our Linear strategy provides a better cross-noise trade-off, remaining competitive at $70\%$ ($11.47\%$) while preventing degradation at $50\%$ ($12.95\%$), making it a more suitable strategy for determining the candidate-set size.

\input{table/ablation_dual_source}

\input{table/ablation_pll_nl_sntd}

\textit{Ablation Study of Dual-Source Partial Label Construction.}
The contribution of different evidence sources used to construct the partial label candidate set is evaluated in Table~\ref{tab:ablation_dual_source}. 
Utilizing EMA-Pred (semantic consensus) as the sole evidence source yields a baseline mIoU of $10.86\%$. 
In contrast, employing only EMA-Proto (feature-prototype similarity) improves semantic discrimination to $11.31\%$ mIoU, which suggests that structural constraints are vital for addressing the challenges of sparse 3D space. 
Notably, the fusion of both sources achieves the peak performance of $11.47\%$ mIoU. 
This result verifies that the semantic evidence from the teacher model and the structural evidence from the prototype space are highly complementary, effectively preventing the omission of ground-truth labels during the candidate construction process.

\textit{Ablation Study of Key Robust Learning Components.}
At $70\%$ noise (Table~\ref{tab:ablation_pll_nl_sntd}), the supervised baseline yields only $10.73\%$ mIoU.
The introduction of partial label learning for disambiguation significantly boosts performance, raising the IoU from $33.39\%$ to $35.57\%$, which demonstrates that relaxing hard targets into a candidate set effectively mitigates noise overfitting. 
The addition of negative learning for noise suppression further improves geometric completeness to $36.07\%$ IoU by explicitly penalizing categories identified as unreliable. 
Finally, incorporating self-not-true distillation for distribution-level regularization yields the highest $11.47\%$ mIoU. 
Crucially, these results reveal that robustness arises not from stronger penalization alone, but from restricting the feasible semantic hypothesis space through dual-source partial labeling, supporting \textit{DPR-Occ}'s underlying principle: sparse 3D learning benefits more significantly from search-space regularization than from global loss rebalancing.

\subsection{Qualitative Analysis}
\begin{figure}[!t]
  \centering
  \includegraphics[width=\linewidth]{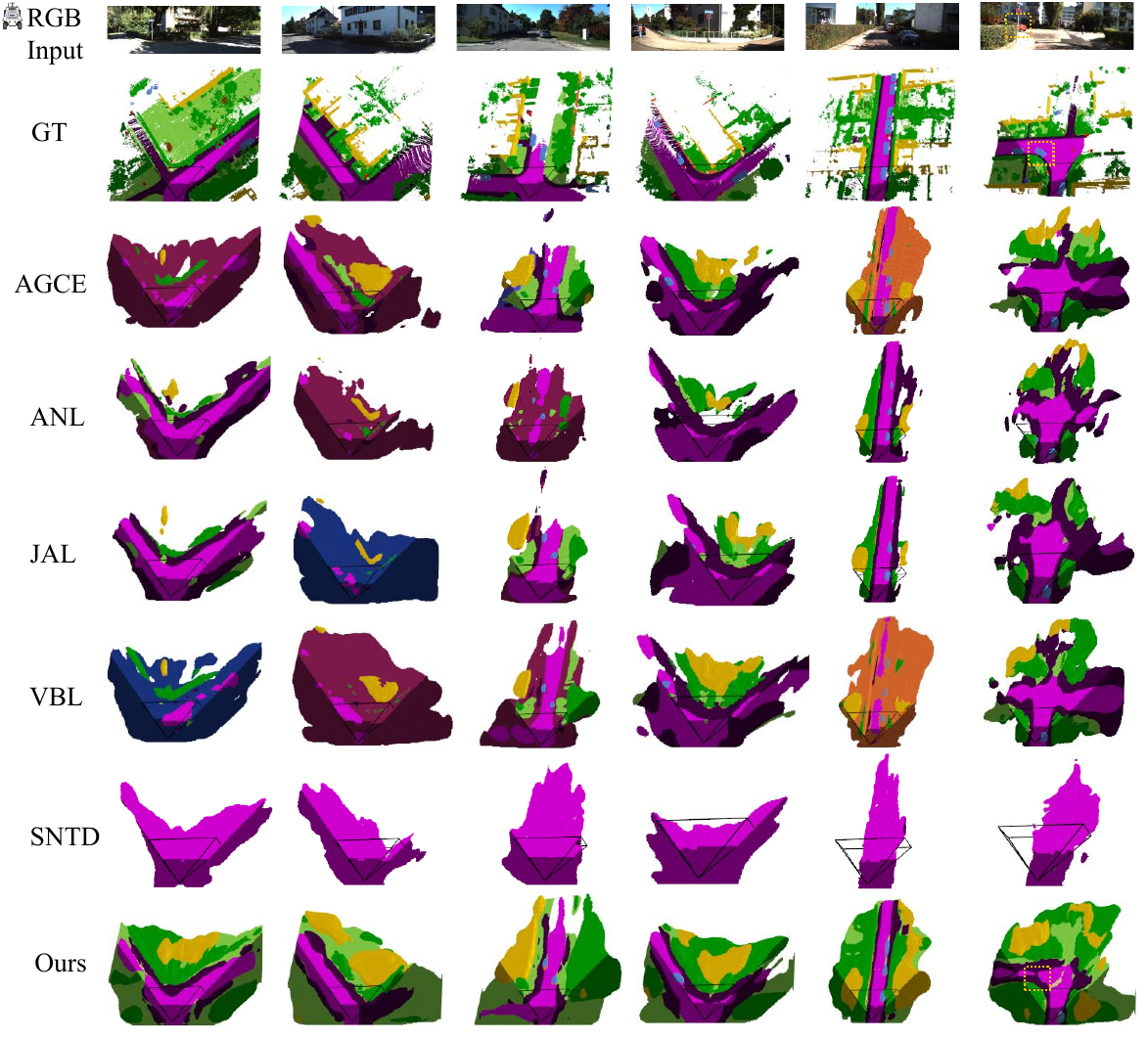}
  \vskip-1ex
  \caption{Qualitative results under $90\%$ asymmetric noise on \textit{OccNL} benchmark. Compared to collapsing baselines, \textit{DPR-Occ} preserves structural integrity and reliable semantics. The final column shows a failure case (yellow box), where \textit{DPR-Occ} still reconstructs basic road and vegetation geometry despite semantic misclassification.}
  \label{fig:occnl_qualitative}
  \vskip-3ex
\end{figure}
As shown in Fig.~\ref{fig:occnl_qualitative}, under extreme $90\%$ asymmetric noise, baselines~\cite{zhou2023asymmetric,ye2024active,wang2025joint,wang2025variation,lan2025continuous} suffer catastrophic structural collapse, losing geometric boundaries and sparse semantic cues. 
In contrast, \textit{DPR-Occ} (first five columns) maintains high-fidelity geometric completeness and reliable semantics. 
This robustness is critical for autonomous driving. 
In long-range regions, sparse observations can reduce annotation reliability, making voxel labels more prone to confusion between empty space and semantic categories. 
While baseline methods would output fragmented predictions that may lead to dangerous phantom braking or collisions, \textit{DPR-Occ} preserves structural integrity, ensuring downstream planning modules receive safe, reliable environmental representations despite severe annotation corruption.

\textit{Failure Case Analysis:} 
Due to the highly ambiguous appearance and confusion with the surrounding context, a distant car is misclassified as vegetation. However, unlike the structural collapse of baselines, \textit{DPR-Occ} still reconstructs the essential surroundings, including most roads and vegetation, providing a safer fallback for navigation.

%% file: Contents/alg/alg_occnl.tex
\begin{algorithm}[H]
\caption{Our proposed \textit{DPR-Occ} algorithm}
\label{alg:occnl}
\textbf{Input:} $\tilde{D}_{\text{train}}$, Student $\theta$, EMA $\theta^{ema}$, epochs $E_w, E_{total}$. \\
\textbf{Output:} Robust Student $\theta$ and stable Teacher $\theta^{ema}$.

\begin{algorithmic}[1]
\For{epoch $e = 1, \dots, E_{total}$}
    \If{$e \le E_w$}
        \State Update $\theta$ by optimizing $\mathcal{L}_{base}$ using noisy $\tilde{Y}$.
    \Else
        \State Update $K_e$~\eqref{eq:dynamic_k} and Calc. Similarity $S$~\eqref{eq:similarity}.
        \State Construct Candidate Set $\mathcal{PL}$~\eqref{eq:union_fusion} and $\mathcal{NL}$.
        \State Calc. $\mathcal{L}_{PLL}$~\eqref{eq:loss_pll}, $\mathcal{L}_{NL}$~\eqref{eq:loss_nl}, $\mathcal{L}_{SNTD}$~\eqref{eq:loss_sntd}.
        \State Update $\theta$ by optimizing $\mathcal{L}_{total}$~\eqref{eq:loss_total}.
    \EndIf
    \State Update EMA $\theta^{ema}$~\eqref{eq:ema_update}.
\EndFor
\end{algorithmic}
\vspace{-2pt}
\end{algorithm}

%% file: table/protoocc_skitti_tail_val.tex
\begin{table*}[t]
\newcommand{\clsname}[2]{%
  \rotatebox{90}{%
    \hspace{-6pt}\textcolor{#2}{$\blacksquare$}\hspace{-6pt}%
    \begin{tabular}{l}#1\end{tabular}%
  }%
}

\centering
\caption{Comparison on \textit{OccNL} benchmark under varying levels of dynamic object trailing noise (Mild, Moderate, and Severe).}
\label{tab:protoocc_skitti_tail_val}
\vskip-1ex
\small
\resizebox{\linewidth}{!}{
\begin{tabular}{c|l|cc|ccccccccccccccccccc}
\toprule
\makecell{Noise \\ Level} &
Method &
{\rotatebox{90}{IoU}} &
{\rotatebox{90}{mIoU}} &
\clsname{road}{road} &
\clsname{sidewalk}{sidewalk} &
\clsname{parking}{parking} &
\clsname{other-grnd.}{otherground} &
\clsname{building}{building} &
\clsname{car}{car} &
\clsname{truck}{truck} &
\clsname{bicycle}{bicycle} &
\clsname{motorcycle}{motorcycle} &
\clsname{other-veh.}{othervehicle} &
\clsname{vegetation}{vegetation} &
\clsname{trunk}{trunk} &
\clsname{terrain}{terrain} &
\clsname{person}{person} &
\clsname{bicyclist}{bicyclist} &
\clsname{motorcyclist}{motorcyclist} &
\clsname{fence}{fence} &
\clsname{pole}{pole} &
\clsname{traf.-sign}{trafficsign} \\
\midrule

\multirow{6}{*}{Mild}
& AGCE & 37.25 & 13.04 & 58.32 & 28.49 & 21.91 & 0.80 & 15.92 & 27.31 & 0.00 & 1.85 & 4.28 & 13.58 & 19.16 & 3.54 & 32.36 & 5.04 & 0.00 & 0.00 & 7.67 & 4.76 & 2.76 \\
& ANL & 36.93 & 13.02 & 58.63 & 28.03 & 18.58 & 0.24 & 15.84 & 26.05 & 7.79 & 3.46 & 3.69 & 9.64 & 19.22 & 3.97 & 32.31 & 5.12 & 0.06 & 0.00 & 7.37 & 4.72 & 2.66 \\
& JAL & 37.18 & 13.14 & 58.63 & 28.61 & 20.96 & 0.54 & 16.54 & 27.40 & 4.93 & 1.76 & 4.14 & 11.60 & 19.58 & 3.62 & 31.51 & 5.50 & 0.00 & 0.00 & 7.12 & 4.76 & 2.49 \\
& VBL & 37.21 & 12.92 & 59.25 & 29.19 & 23.23 & 0.12 & 16.01 & 26.97 & 0.00 & 0.88 & 2.68 & 12.82 & 19.12 & 3.47 & 31.43 & 5.16 & 0.00 & 0.00 & 7.82 & 4.61 & 2.65 \\
& SNTD & 36.72 & 13.42 & 58.73 & 29.64 & 21.20 & 0.11 & 15.10 & 27.51 & 16.07 & 1.60 & 2.44 & 11.08 & 17.79 & 2.27 & 32.58 & 6.63 & 0.00 & 0.00 & 7.71 & 2.44 & 2.00 \\
\rowcolor[gray]{.9} \multicolumn{1}{c|}{\cellcolor{white}} & Ours & 34.94 & 13.82 & 58.48 & 27.84 & 21.33 & 0.02 & 15.39 & 24.40 & 13.91 & 3.82 & 5.38 & 13.63 & 18.92 & 3.62 & 32.18 & 4.46 & 4.36 & 0.00 & 7.73 & 4.48 & 2.67 \\
\midrule

\multirow{6}{*}{Moderate}
& AGCE & 37.18 & 12.00 & 58.97 & 27.99 & 21.32 & 0.03 & 15.80 & 26.34 & 0.00 & 2.50 & 2.76 & 0.00 & 18.92 & 3.72 & 31.96 & 4.54 & 0.00 & 0.00 & 7.01 & 4.42 & 1.63 \\
& ANL & 37.18 & 13.04 & 58.46 & 28.46 & 23.24 & 0.88 & 16.09 & 25.95 & 6.48 & 2.14 & 4.11 & 10.58 & 19.12 & 3.19 & 31.27 & 4.20 & 0.00 & 0.00 & 7.31 & 4.45 & 1.82 \\
& JAL & 37.28 & 12.96 & 59.77 & 29.29 & 21.62 & 0.14 & 15.70 & 26.41 & 5.23 & 0.91 & 3.00 & 11.90 & 19.21 & 3.45 & 31.56 & 3.71 & 0.00 & 0.00 & 8.01 & 4.24 & 2.17 \\
& VBL & 37.41 & 13.15 & 59.60 & 27.84 & 22.50 & 0.92 & 15.73 & 26.70 & 6.70 & 1.35 & 2.10 & 13.06 & 19.27 & 3.66 & 31.25 & 4.12 & 0.00 & 0.00 & 7.63 & 4.65 & 2.85 \\
& SNTD & 35.76 & 13.07 & 58.87 & 30.33 & 19.14 & 0.00 & 15.20 & 26.78 & 14.38 & 1.77 & 2.74 & 10.94 & 17.61 & 1.15 & 30.56 & 5.06 & 5.92 & 0.00 & 6.49 & 0.90 & 0.40 \\
\rowcolor[gray]{.9} \multicolumn{1}{c|}{\cellcolor{white}} & Ours & 35.73 & 13.42 & 58.68 & 27.65 & 22.68 & 0.00 & 15.52 & 24.90 & 15.34 & 0.76 & 3.30 & 13.86 & 18.89 & 3.74 & 31.77 & 3.34 & 0.79 & 0.00 & 6.99 & 4.26 & 2.56 \\
\midrule

\multirow{6}{*}{Severe}
& AGCE & 36.95 & 11.58 & 59.19 & 28.41 & 21.10 & 0.12 & 16.05 & 22.13 & 0.00 & 2.01 & 0.00 & 0.00 & 19.04 & 3.41 & 32.09 & 3.21 & 0.00 & 0.00 & 7.64 & 4.74 & 0.97 \\
& ANL & 37.21 & 13.27 & 59.18 & 28.55 & 20.80 & 0.14 & 16.15 & 23.78 & 9.35 & 1.49 & 3.65 & 12.04 & 18.96 & 3.76 & 31.95 & 3.51 & 3.63 & 0.00 & 7.79 & 4.62 & 2.77 \\
& JAL & 37.32 & 12.87 & 59.17 & 28.77 & 22.88 & 0.57 & 16.09 & 25.77 & 2.45 & 0.20 & 2.86 & 10.74 & 19.36 & 3.73 & 32.65 & 3.63 & 1.79 & 0.00 & 7.30 & 4.16 & 2.36 \\
& VBL & 37.27 & 12.64 & 59.50 & 28.66 & 22.24 & 0.01 & 16.36 & 22.74 & 0.00 & 0.95 & 3.19 & 12.37 & 18.91 & 3.57 & 32.91 & 2.63 & 1.27 & 0.00 & 7.24 & 4.53 & 3.02 \\
& SNTD & 35.73 & 12.67 & 59.39 & 30.41 & 21.92 & 0.00 & 13.92 & 25.16 & 4.38 & 3.29 & 1.68 & 11.04 & 17.79 & 1.05 & 31.43 & 4.52 & 3.88 & 0.00 & 7.17 & 1.60 & 2.10 \\
\rowcolor[gray]{.9} \multicolumn{1}{c|}{\cellcolor{white}} & Ours & 35.70 & 13.67 & 58.93 & 28.11 & 20.35 & 0.48 & 15.38 & 23.41 & 14.99 & 3.83 & 3.85 & 12.88 & 19.48 & 3.93 & 32.35 & 3.94 & 3.95 & 0.00 & 7.13 & 4.06 & 2.63 \\
\bottomrule
\end{tabular}
}
\vskip-3ex
\end{table*}

%% file: table/ablation_Ew.tex
\begin{table}[!t]
  \centering
  \resizebox{\linewidth}{!}{
    \begin{minipage}[t]{0.35\linewidth}
      \centering
      \caption{Ablation of $E_{w}$ at $70\%$ noise.}
      \label{tab:ablation_ew}
      \vskip-1ex
      \begin{tabular}{@{}ccc@{}} %
        \toprule
        $E_{w}$ & \makecell{IoU\\(\%)} & \makecell{mIoU\\(\%)}\\
        \midrule
        6  & 31.71 & 11.02 \\
        8  & 33.15 & 11.55 \\
        10 & 33.80 & 11.44 \\
        12 & 35.86 & 11.47 \\
        14 & 35.96 & 11.28 \\
        16 & 33.65 & 11.03 \\
        18 & 28.62 & 9.27 \\
        \bottomrule
      \end{tabular}
    \end{minipage}
    \hfill %
    \begin{minipage}[t]{0.65\linewidth}
      \centering
      \captionsetup{margin={12pt, 0pt}}
      \caption{Ablation of Dynamic-$K$ \\Scheduling at $50\%$ and $70\%$ noise.}
      \label{tab:ablation_dynamic_k}
      \vskip-1ex
      \begin{tabular}{@{}cccc@{}}
        \toprule
        \makecell{Noise \\ Rate(\%)}& Strategy & \makecell{IoU\\(\%)} & \makecell{mIoU\\(\%)}\\
        \midrule
        70 & Linear (\ref{eq:dynamic_k}) & 35.86 & 11.47 \\
        70 & Fixed 2 & 29.43 & 10.59 \\
        70 & Fixed 5 & 36.43 & 11.15 \\
        70 & Random  & 35.47 & 11.43 \\
        70 & Fixed 9 & 34.26 & 11.56 \\
        50 & Fixed 9 & 32.21 & 11.48 \\
        50 & Linear (\ref{eq:dynamic_k}) & 35.78 & 12.95 \\
        \bottomrule
      \end{tabular}
    \end{minipage}
  }
  \vskip-3ex
\end{table}

%% file: table/ablation_dual_source.tex
\begin{table}[t]
\centering
\caption{Ablation study of dual-source evidence for candidate set construction at $70\%$ noise. 
\textbf{EMA-Pred}: Semantic consensus from the EMA teacher; 
\textbf{EMA-Proto}: Representation-structure evidence from feature-prototype similarity.}
\label{tab:ablation_dual_source}
\vskip-1ex
\small
\begin{tabular}{cccc} 
\toprule
EMA-Pred & EMA-Proto    & IoU (\%)      & mIoU (\%)       \\ \midrule
\checkmark &            & 35.88          & 10.86          \\
           & \checkmark & 36.24          & 11.31          \\ \midrule
\checkmark & \checkmark & 35.86          & 11.47          \\ \bottomrule
\end{tabular}
\vskip-3ex
\end{table}

%% file: table/ablation_pll_nl_sntd.tex
\begin{table}[t]
\centering
\caption{Ablation study of the key robust learning components at $70\%$ noise. \textbf{Disamb.}: Disambiguation via Partial Label Learning; \textbf{Supp.}: Noise Suppression via Negative Learning; \textbf{Regu.}: Distribution Regularization via EMA-guided SNTD.}
\label{tab:ablation_pll_nl_sntd}
\vskip-1ex
\small
\setlength{\tabcolsep}{6.5pt}
\begin{tabular}{cccccc}
\toprule
Standard & Disamb & Supp & Regu & IoU (\%) & mIoU (\%) \\ \midrule
\checkmark &           &           &           & 33.39 & 10.73 \\
\checkmark & \checkmark &           &           & 35.57 & 11.42 \\
\checkmark & \checkmark & \checkmark &           & 36.07 & 11.31 \\
\checkmark & \checkmark & \checkmark & \checkmark & 35.86 & 11.47 \\ \bottomrule
\end{tabular}
\vskip-2ex
\end{table}

%% file: Contents/Conclusion.tex
We introduce \textit{OccNL}, the first benchmark for systematically studying 3D semantic occupancy under occupancy-asymmetric and dynamic trailing label noise. 
Our analysis reveals a key vulnerability in sparse voxel learning: semantic degradation occurs before structural collapse, and image-domain robust objectives can cause minority semantic extinction under severe corruption.
To address this vulnerability, we propose \textit{DPR-Occ}, a noise-robust framework based on dual-source partial label reasoning. 
By integrating temporal model memory and representation-level structural affinity, \textit{DPR-Occ} constrains the feasible semantic hypothesis space rather than relying solely on loss reweighting. 
This structured regularization effectively delays representation collapse, preserving both geometric integrity and sparse dynamic semantics under extreme label corruption.
Ultimately, our findings highlight that robust 3D perception differs fundamentally from image-level learning, relying more on hypothesis-space control and structural consistency than on stronger penalization alone.
Future work will integrate noise-robust occupancy into closed-loop autonomous systems and jointly model long-tailed dynamics and spatiotemporal artifacts for reliable real-world deployment.

%% file: main.bbl
\begin{thebibliography}{10}
\providecommand{\url}[1]{#1}
\csname url@samestyle\endcsname
\providecommand{\newblock}{\relax}
\providecommand{\bibinfo}[2]{#2}
\providecommand{\BIBentrySTDinterwordspacing}{\spaceskip=0pt\relax}
\providecommand{\BIBentryALTinterwordstretchfactor}{4}
\providecommand{\BIBentryALTinterwordspacing}{\spaceskip=\fontdimen2\font plus
\BIBentryALTinterwordstretchfactor\fontdimen3\font minus \fontdimen4\font\relax}
\providecommand{\BIBforeignlanguage}[2]{{%
\expandafter\ifx\csname l@#1\endcsname\relax
\typeout{** WARNING: IEEEtran.bst: No hyphenation pattern has been}%
\typeout{** loaded for the language `#1'. Using the pattern for}%
\typeout{** the default language instead.}%
\else
\language=\csname l@#1\endcsname
\fi
#2}}
\providecommand{\BIBdecl}{\relax}
\BIBdecl

\bibitem{li2025voxdet}
W.~Li, Z.~Yu, and A.~Alahi, ``{VoxDet:} {Rethinking} {3D} semantic occupancy prediction as dense object detection,'' in \emph{Proc. NeurIPS}, 2025.

\bibitem{wang2025uniocc}
Y.~Wang \emph{et~al.}, ``{UniOcc:} {A} unified benchmark for occupancy forecasting and prediction in autonomous driving,'' in \emph{Proc. ICCV}, 2025, pp. 25\,560--25\,570.

\bibitem{marcuzzi2025sfmocc}
R.~Marcuzzi \emph{et~al.}, ``{SfmOcc:} {Vision-based} {3D} semantic occupancy prediction in urban environments,'' \emph{IEEE Robotics and Automation Letters}, vol.~10, no.~5, pp. 5074--5081, 2025.

\bibitem{kim2025vpocc}
J.~Kim, J.~Lee, U.~Shin, J.~Oh, and K.~Joo, ``{VPOcc:} {Exploiting} vanishing point for {3D} semantic occupancy prediction,'' in \emph{Proc. IROS}, 2025, pp. 4307--4314.

\bibitem{zuo2025quadricformer}
S.~Zuo, W.~Zheng, X.~Han, L.~Yang, Y.~Pan, and J.~Lu, ``{QuadricFormer:} {Scene} as superquadrics for {3D} semantic occupancy prediction,'' in \emph{Proc. NeurIPS}, 2025.

\bibitem{reed2025online}
A.~Reed, L.~Achey, B.~Crowe, B.~Hayes, and C.~Heckman, ``Online diffusion-based {3D} occupancy prediction at the frontier with probabilistic map reconciliation,'' in \emph{Proc. ICRA}, 2025, pp. 2846--2852.

\bibitem{shi2025h3o}
Y.~Shi, H.~Cai, A.~Ansari, and F.~Porikli, ``{H3O:} {Hyper-efficient} {3D} occupancy prediction with heterogeneous supervision,'' in \emph{Proc. ICRA}, 2025, pp. 4869--4876.

\bibitem{su2024alpha}
S.~Su, N.~Chen, C.~Lin, F.~Juefei-Xu, C.~Feng, and F.~Miao, ``$\alpha$-{OCC}: {Uncertainty-aware} camera-based {3D} semantic scene completion,'' \emph{arXiv preprint arXiv:2406.11021}, 2024.

\bibitem{chen2025semantic}
D.~Chen \emph{et~al.}, ``Semantic causality-aware vision-based {3D} occupancy prediction,'' in \emph{Proc. ICCV}, 2025, pp. 24\,878--24\,888.

\bibitem{kim2025protoocc}
J.~Kim, C.~Kang, D.~Lee, S.~Choi, and J.~W. Choi, ``{ProtoOcc:} {Accurate}, efficient {3D} occupancy prediction using dual branch encoder-prototype query decoder,'' in \emph{Proc. AAAI}, 2025, pp. 4284--4292.

\bibitem{zhou2025autoocc}
X.~Zhou, J.~Wang, Y.~Wang, Y.~Wei, N.~Dong, and M.-H. Yang, ``{AutoOcc:} {Automatic} open-ended semantic occupancy annotation via vision-language guided gaussian splatting,'' in \emph{Proc. ICCV}, 2025, pp. 3367--3377.

\bibitem{behley2019semantickitti}
J.~Behley \emph{et~al.}, ``{SemanticKITTI:} {A} dataset for semantic scene understanding of {LiDAR} sequences,'' in \emph{Proc. ICCV}, 2019, pp. 9296--9306.

\bibitem{zhou2023asymmetric}
X.~Zhou, X.~Liu, D.~Zhai, J.~Jiang, and X.~Ji, ``Asymmetric loss functions for noise-tolerant learning: Theory and applications,'' \emph{IEEE Transactions on Pattern Analysis and Machine Intelligence}, vol.~45, no.~7, pp. 8094--8109, 2023.

\bibitem{wang2025variation}
J.~Wang \emph{et~al.}, ``Variation-bounded loss for noise-tolerant learning,'' in \emph{Proc. AAAI}, 2026, pp. 26\,251--26\,259.

\bibitem{ye2024active}
X.~Ye, Y.~Wu, Y.~Wang, X.~Li, W.~Zhang, and Y.~Chen, ``Active negative loss: A robust framework for learning with noisy labels,'' \emph{arXiv preprint arXiv:2412.02373}, 2024.

\bibitem{wang2025joint}
J.~Wang \emph{et~al.}, ``Joint asymmetric loss for learning with noisy labels,'' in \emph{Proc. ICCV}, 2025, pp. 1947--1956.

\bibitem{lan2025continuous}
L.~Lan, J.~Wang, X.~Wu, B.~Han, and X.~Liu, ``Continuous review and timely correction: Enhancing the resistance to noisy labels via self-not-true and class-wise distillation,'' \emph{IEEE Transactions on Pattern Analysis and Machine Intelligence}, vol.~48, no.~5, pp. 5165--5179, 2025.

\bibitem{heidrich2025occuq}
S.~Heidrich, T.~Beemelmanns, A.~Nekrasov, B.~Leibe, and L.~Eckstein, ``{OCCUQ:} {Exploring} efficient uncertainty quantification for {3D} occupancy prediction,'' in \emph{Proc. ICRA}, 2025, pp. 1--8.

\bibitem{wang2025reliable}
S.~Wang \emph{et~al.}, ``Reliable and calibrated semantic occupancy prediction by hybrid uncertainty learning,'' in \emph{Proc. IJCAI}, 2025, pp. 1973--1981.

\bibitem{chen2025particle}
G.~Chen, Z.~Wang, W.~Dong, and J.~Alonso-Mora, ``Particle-based instance-aware semantic occupancy mapping in dynamic environments,'' \emph{IEEE Transactions on Robotics}, vol.~41, pp. 1155--1171, 2025.

\bibitem{deng2022_hd_ccsom}
Y.~Deng, M.~Wang, Y.~Yang, and Y.~Yue, ``{HD-CCSOM:} {Hierarchical} and dense collaborative continuous semantic occupancy mapping through label diffusion,'' in \emph{Proc. IROS}, 2022, pp. 2417--2422.

\bibitem{liu2025sparse_annotation}
Z.~Liu \emph{et~al.}, ``Sparse annotation, dense supervision: Unleashing self-training power for occupancy prediction with {2D} labels,'' \emph{IEEE Robotics and Automation Letters}, vol.~11, no.~1, pp. 418--425, 2026.

\bibitem{li2025enhancing_unlabeled}
R.~Li \emph{et~al.}, ``Enhancing generalizability via utilization of unlabeled data for occupancy perception,'' in \emph{Proc. AAAI}, 2025, pp. 4896--4904.

\bibitem{yu2025language_driven}
Z.~Yu \emph{et~al.}, ``Language driven occupancy prediction,'' in \emph{Proc. ICCV}, 2025, pp. 7548--7558.

\bibitem{zhu2024nucraft}
B.~Zhu, Z.~Wang, and H.~Li, ``{nuCraft:} {Crafting} high resolution {3D} semantic occupancy for unified {3D} scene understanding,'' in \emph{Proc. ECCV}, 2024, pp. 125--141.

\bibitem{jiang2023knowledge_distillation}
R.~Jiang, Y.~Yan, J.-H. Xue, S.~Chen, N.~Wang, and H.~Wang, ``Knowledge distillation meets label noise learning: {Ambiguity-guided} mutual label refinery,'' \emph{IEEE Transactions on Neural Networks and Learning Systems}, vol.~36, no.~1, pp. 939--952, 2025.

\bibitem{bai2025robust_noisy_label_learning}
S.~Bai, S.~Zhou, Z.~Qin, L.~Wang, and N.~Zheng, ``Robust noisy label learning via two-stream sample distillation,'' \emph{IEEE Transactions on Multimedia}, vol.~27, pp. 9072--9084, 2025.

\bibitem{das2023understanding_self_distillation}
R.~Das and S.~Sanghavi, ``Understanding self-distillation in the presence of label noise,'' in \emph{Proc. ICML}, 2023, pp. 7102--7140.

\bibitem{lu2024federated}
Y.~Lu \emph{et~al.}, ``Federated learning with extremely noisy clients via negative distillation,'' in \emph{Proc. AAAI}, 2024, pp. 14\,184--14\,192.

\bibitem{tarvainen2017mean}
A.~Tarvainen and H.~Valpola, ``Mean teachers are better role models: Weight-averaged consistency targets improve semi-supervised deep learning results,'' in \emph{Proc. NeurIPS}, 2017, pp. 1195--1204.

\bibitem{gao2023semi}
H.~Gao \emph{et~al.}, ``From semi-supervised to omni-supervised room layout estimation using point clouds,'' in \emph{Proc. ICRA}, 2023, pp. 2803--2810.

\bibitem{tang2024sparseocc}
P.~Tang \emph{et~al.}, ``{SparseOcc:} {Rethinking} sparse latent representation for vision-based semantic occupancy prediction,'' in \emph{Proc. CVPR}, 2024, pp. 15\,035--15\,044.

\bibitem{xia2023scpnet}
Z.~Xia \emph{et~al.}, ``{SCPNet:} {Semantic} scene completion on point cloud,'' in \emph{Proc. CVPR}, 2023, pp. 17\,642--17\,651.

\bibitem{arpit2017closer}
D.~Arpit \emph{et~al.}, ``A closer look at memorization in deep networks,'' in \emph{Proc. ICML}, 2017, pp. 233--242.

\bibitem{he2016deep}
K.~He, X.~Zhang, S.~Ren, and J.~Sun, ``Deep residual learning for image recognition,'' in \emph{Proc. CVPR}, 2016, pp. 770--778.

\end{thebibliography}
